\documentclass[letterpaper, 10 pt, conference]{ieeeconf}  % Comment this line out if you need a4paper

\IEEEoverridecommandlockouts                              % This command is only needed if 
\let\labelindent\relax
\usepackage{float}
\usepackage[caption = true]{subfig}
\usepackage{url}
\usepackage{bm,bbold}
\usepackage{textcomp}
\usepackage{indentfirst}
\usepackage{xcolor}
\usepackage{multirow}
\usepackage{enumitem}
\usepackage{graphicx}
\usepackage{amsmath}
\usepackage{amssymb}
\usepackage{booktabs}

\title{Shadow Transfer: Single Image Relighting For Urban Road Scenes}
\author{Alexandra Carlson$^{1}$, Ram Vasudevan$^{2}$ and Matthew Johnson-Roberson$^{3}$
\thanks{$^{1}$A. Carlson is with the Robotics Institute, University of Michigan, Ann Arbor, MI 48109, USA {\tt\small \{askc\}@umich.edu}}
\thanks{$^{2}$R. Vasudevan is with the Department of Mechanical Engineering, University of Michigan, Ann Arbor, MI 48109, USA {\tt\small ramv@umich.edu}}
\thanks{$^{3}$M. Johnson-Roberson is with the Department of Naval Architecture and Marine Engineering, University of Michigan, Ann Arbor, MI 48109, USA {\tt\small mattjr@umich.edu}}
}
\begin{document}

\maketitle
%%%%%%%%%%%%%%%%%%%%%%%%%%%%%%%%%%%%%%%%%%%%%%%%%%%%%%%%%%%%%%%%%%%%
%%%%%%%%%%%%%%%%%%%%%%%%%%%%%%%%%% ABSTRACT %%%%%%%%%%%%%%%%%%%%%%%%%%%%%%%%%%%%%%
%%%%%%%%%%%%%%%%%%%%%%%%%%%%%%%%%%%%%%%%%%%%%%%%%%%%%%%%%%%%%%%%%%%%
\begin{abstract}
\label{sec:abstract}
%Illumination-based image artifacts
Illumination effects in images, specifically cast shadows and shading, have been shown  to decrease  the performance of deep neural networks on a large number of vision-based detection, recognition and segmentation tasks in urban driving scenes. A key factor that contributes to this performance gap is the lack of `time-of-day' diversity within real, labeled datasets. 
There have been impressive advances in the realm of image to image translation in transferring previously unseen visual effects into a dataset, specifically in day to night translation. However, it is not easy to constrain what visual effects, let alone illumination effects, are transferred from one dataset to another during the training process. 
To address this problem, we propose deep learning framework, called Shadow Transfer, that can relight complex outdoor scenes by transferring realistic shadow, shading, and other lighting effects onto a single image. The novelty of the proposed framework is that it is both self-supervised, and is designed to operate on sensor and label information that is easily available in autonomous vehicle datasets. 
We show the effectiveness of this method on both synthetic and real datasets, and we provide experiments that demonstrate that the proposed method produces images of higher visual quality than state of the art image to image translation methods. % without the burden of capturing specific domains. 
\end{abstract}
%
%% Keywords appear just beneath the abstract. Use only for final RAL version.  
%\begin{IEEEkeywords}
%Deep Learning in Robotics and Automation; Visual Learning; Computer Vision for Other Robotic %Applications; Simulation and Animation
%\end{IEEEkeywords}

%%%%%%%%%%%%%%%%%%%%%%%%%%%%%%%%%%%
%%%%%%%%% PAPER MAIN BODY %%%%%%%%%
%%%%%%%%%%%%%%%%%%%%%%%%%%%%%%%%%%%

%% motivate shadows as failure mode figure
\begin{figure}
\begin{center}
\includegraphics[width=0.8\linewidth]{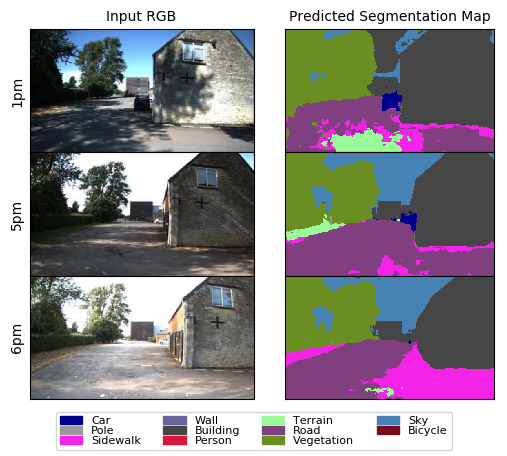}
\caption{For different times of day (each row), while the scene geometry remains consistent, there is are significant changes in the appearance of the scene, both locally (e.g., object shadows) as well as globally (global color temperature and scene brightness). \vspace{-6mm}}
\label{fig:introfig}
\end{center}
\end{figure}
%
%%%%%%%%% INTRO TEXT %%%%%%%%%
\section{Introduction}
\label{sec:intro}
%% include the intro text
%%%%%%%%%%%%%%%%%%%%
%%% INTRODUCTION %%%
%%%%%%%%%%%%%%%%%%%%
%Shadows provide useful cues about the scene including object shapes, light sources and illumination conditions, camera parameters, and scene geometry. On the other hand, the presence of shadows in images creates difficulties for many computer vision tasks from image segmentation to object detection and tracking. In all cases, being able to automatically detect shadows, and subsequently remove them or reason about their shapes and sizes would usually be beneficial.

In outdoor road environments, the appearance of a salient object, such as a car or pedestrian, is highly dependent upon the illumination of the scene. Adjusting the time of day can induce significant changes to the scene's appearance despite the fact that its underlying structure and material properties remain the same. Illumination conditions at different times of day can alter scene and object appearance in two primary ways. First, the location and angle of incoming sunlight interacts with the physical structure of the scene to induce different distributions of cast shadows and shading within the scene. Second, the properties of sunlight can interact with the imaging sensor to alter perceived colors within the scene, for example, differing times of day can have different color temperatures and/or overall brightness. 
These ``time of day" visual features have been shown to decrease performance of deep neural networks (DNNs) for a variety of vision-based detection, recognition, and segmentation tasks in outdoor driving scenarios~\cite{alshammari2018impact, bahri2018online, failingtolearn}.
In Figure \ref{fig:introfig}, the state-of-the-art DeepLab~\cite{chen2017deeplab} semantic segmentation framework, trained on Cityscapes~\cite{Cityscapes}, is tested on images of the same scene under different times of day taken from the Oxford Robot dataset~\cite{maddern20171}. The segmentation performance fluctuates drastically between images captured within hours of each other, particularly in the image regions where the distribution of shadows and brightness change significantly. 
The fact that sun position/time of day is source of prediction error for DNNs suggests that the network's learned feature representations are dependent upon the illumination of the captured scene, which in turns indicates that real datasets do not have enough diversity in lighting conditions to allow the network to become invariant to such changes.
Unfortunately, it is too costly to collect and label real datasets that capture a uniform distribution of lighting effects over the course of a day. While it is possible to generate synthetic datasets that capture a uniform representation of illumination effects like shadows using rendering pipelines and gaming engines~\cite{Richter_2016_ECCV,johnson2016driving}, they have a significant and undesirable domain gap from real data~\cite{johnson2016driving}.
In contrast, we pose that, in order to improve the robustness of these neural network methods to varying illumination conditions, we need to more accurately and reliably model illumination changes in real, noisy images.  
While there has been impressive advances in image relighting within the graphics and vision communities~\cite{philip2019multi,xu2018deep}, these methods rely on having knowledge of material properties of objects in the scene, or having multiple views of the scene under the same illumination condition, neither of which is possible with driving datasets.
The method proposed in this paper, which we dub ``Shadow Transfer", leverages the incredible success of image to image translation models to learn an illumination model via a deep neural encoder-decoder framework that operates upon input that is easily obtained from a car-mounted RGB camera. Furthermore, it is designed to be self-supervised, removing the need for labeling illumination features in images, like shadows, brightness or global color temperature.
%Given a coarse depth and semantic segmentation maps, and the corresponding RGB image, the Shadow Transfer framework can be used to apply realistic lighting changes to the RGB image.
%Given the current sun position, a coarse depth representation, coarse semantic segmentation, and corresponding RGB image as input, the Shadow Transfer framework relights the RGB image in two stages: first, it learns the spatial relationship between scene geometry and light source so it can transfer accurate shadows and shading, and then it learns an illumination-based colorization model that controls the changes in color temperature, exposure and brightness that is present in the real world.
%We anticipate that the proposed model is the first step in an 
%effort to add useful visual manipulations to images by applying our proposed relighting pipeline to domain randomization for any vision task.
To our knowledge, this is the first attempt at specifically relighting driving datasets by transferring lighting features. We anticipate that the proposed model, by adding in realistic illumination artifacts into images, is the first step towards understanding and eliminating the prediction error of detection and segmentation algorithms that results from changing lighting conditions.

%%%%%%%%%%%%%%%%%%%%%%%%%%%%%%
%
%%%%%%%%% BACKGROUND AND RELATED WORKS TEXT %%%%%%%%%
\section{Related work}
\label{sec:background}
%% include the background text
%%%%%%%%%%%%%%%%%%%%%%%%%%%%%%%
%%% PRIOR AND RELATED WORKS %%%
%%%%%%%%%%%%%%%%%%%%%%%%%%%%%%%
%The problem of single image relighting is highly unconstrained because the depth, texture, and albedo of the scene is unknown. Thus, relighting methods require basic assumptions about either scene geometry, materials, or a prior illumination model. 
In this work, the objective is to transfer realistic lighting effects onto an image given a coarse model of scene geometry and a single RGB image. This process requires both removing and re-synthesizing illumination features within the image. 
%For the Shadow Transfer framework, given an input RGB image of a scene and geometric representation in the form of semantic segmentation and depth maps, we want to re-render the illumination artifacts in the scene using intuition from image to image translation methods, shadow modeling methods, and inverse graphics rendering. 
The proposed Shadow Transfer method is based upon deep neural network architectures used for image-to-image translation, image relighting, shadow modeling, outdoor illumination estimation, and inverse graphics rendering. We briefly review the literature in these areas below.
%The majority of this research has either focused on removing the effect of illumination upon the scene, e.g., illumination invariant images and shadow removal, or focused on modeling and extracting the shape and material properties of objects within the scene, e.g., intrinsic image decomposition.
%State of the art illumination modeling methods all employ deep learning techniques. 
%The majority of methods use encoder-decoder frameworks, in which one neural network encodes a scene representation into a compressed form/code, then a separate network decodes this feature code. 
%The goal of the proposed Shadow Transfer method is to transfer lighitng effects onto an image, which requires both removal and synthesis.
%builds upon a subset of these, which we describe below.
%
%%%%%%%%%%%%%%%%%%%%%%%%%%%%%%%%%%%%%%%%%%%%%%%%%%%%%%%%%%%%%%%%%%%%%%%%%%%%%%%%%%%%%%%%%%%%%%%%%
~\vspace{-2mm}
\subsection{Illumination Estimation in images}
In order to change the lighting of an image, it is often necessary to first estimate the current illumination conditions. This problem has been broken down into smaller areas, each focusing on the estimation of a specific illumination cue from an image or sequence of images. We briefly review the relevant methods in the following subsections.

\subsubsection{Inverse graphics and Intrinsic Images for Reflectance and Shading Estimation}
Intrinsic image decomposition and inverse graphics rendering decompose an image into a set of intermediate representations or features that correspond to different physical process in the real world, e.g, normal maps, albedo, shading, and reflectance  maps. These intermediate representations can be sampled to synthesize new unseen images, which allows for image relighting. 
There are drawbacks to these techniques that prevent them from being applied to large scale outdoor scenes. First, these methods make simplifying assumptions about the scene structure to make the reconstruction tractable, and thus are usually applied to scenes that contain similar spatial information, such as faces~\cite{chen2019cyclically, worrall2017interpretable, wang2019adversarial}. 
%Second, they generally assume a lambertian reflectance model for the scene surfaces~\cite{}, which does not hold true in real world environments, especially ones with cars. 
Second, they can require rigid scene priors~\cite{barron2013intrinsic,barron2014shape}, such as lambertian reflection and/or shape models, or require complicated training regimes that require specially curated datasets~\cite{donahue2017semantically,kulkarni2015deep}, none of which generalizes to real world outdoor scenes.
%Lastly, they either require multiple images of the same scene~\cite{philip2019multi}, labeled datasets of albedo/shading that are difficult to collect for outdoor scenes~\cite{}, or require complicated training regimes corresponding to complex synthetic datasets~\cite{donahue2017semantically,kulkarni2015deep}. 
%%, e.g., for faces, human skin generally is not reflective, but this constraint
%So global illumination modeling is difficult because its hard to accurately capture the physics of the problem, can’t we just have a neural network take care of everything? 
%Inverse graphics rendering and Intrinsic image decomposition methods both attempt to reverse-engineer the physical processes or features that produced a given image.
%Inferring illumination, shading, and materials in a scene are part of the process of inferring an interpretable feature space or intermediate image representation 
%(3D geometry, illumination, materials) such that a graphics renderer could realistically reproduce the observed scene. 
%Intrinsic image decomposition specifically involves decomposing a natural image into a set of images corresponding to different physical causes, e.g, normal maps, albedo, shading maps.
%It is a "bottom up” approach; that is, from images and videos we extract intrinsic features or intrinsic images, which represent physical properties of the scene tied to the pixel grid.  
%
%%%%%%%%%%%%%%%%%%%%%%%%%%%%%%%%%%%%%%%%%%%%%%%%%%%%%%%%%%%%%%%%%%%%%%%%%%%%%%%%%%%%%%%%%%%%%%%%%
\subsubsection{Shadow-specific modeling in Images}
%The accurate recognition of shadow area, i.e., shadow detection, provides useful information about the light sources, illumination conditions, object shapes and geometry information. 
%Shadows can either improve visual perception and understanding of a scene by adding important shape/orientation information about an object or cues about lighting/scene geometry, or they can add confusion, depending on whether they are inlcuded or ignored for a given task. % or taken into account based on the definition of the loss function.
The impact of shadows on scene understanding has motivated significant research in the development of both deep learning and hand-crafted techniques for shadow detection and removal ~\cite{zheng2019distraction, wang2018stacked, qu2017deshadownet, xiao2014shadow}. 
However, the majority of these methods are designed to produce accurate results on high quality images that contain simple scenes, specifically ones that contain only shadows cast upon a textured ground plane with good lighting conditions. They also typically require material and/or shadow labels, which are prohibitively expensive to generate for real datasets. 
The opposite problem to shadow detection and removal, shadow synthesis, has also been explored, but primarily in the context of adding synthetic objects into images ~\cite{zhang2019shadowgan,alhaija2018geometric,lalonde2012estimating}. These methods also require hard to obtain shadow or material labels, or rigid and hand crafted illumination priors that are difficult to generalize to noisy outdoor datasets. These design choices prevent these methods from generalizing to highly complex scenes that are captured in many autonomous vehicle datasets.
%Shadow synthesis techniques can operate on more complex outdoor datasets that contain the 3D objects that cast shadows. However, they also suffer from the same problems as shadow removal and detection techniques; they typically require intermediate image representations that are not readily available for real outdoor images, such as material segmentation or shadow masks. Furthermore, shadow synthesis methods typically can only be applied to one object inserted into the scene, where as relighting an image would require applying realistic shadow effects to all objects within the scene.
%These shadow understanding methods can be used as a starting point for us though, because each method, at a high level, is learning a relationship between some sort of scene geometry/representation and the appearance. 
In contrast, while the proposed method uses a multi-task structure inspired by~\cite{wang2018stacked}, it is able to model both shadows and shading on scene structures without the need for shadow or material labels. 
%
%%%%%%%%%%%%%%%%%%%%%%%%%%%%%%%%%%%%%%%%%%%%%%%%%%%%%%%%%%%%%%%%%%%%%%%%%%%%%%%%%%%%%%%%%%%%%%%%%
\subsubsection{Sun position Estimation and Modeling}
%Global illumination is the combination of a variety of physical effects, such as indirect illumination, soft shadows, specularities, cast shadows, reflections, crepuscular rays, caustics, etc , and how these manifest in the wild (e.g., under different illuminants, scenes with different geometries, materials). It is a necessary precursor/important component to any relighting technique.
%Human can fairly easily estimate the approximate sun orientation from a non-cloudy outdoor image, relying on shadow directions, highlight positions etc. 
%Extracting these high-level illumination cues/factors from a single image is non-trivial and highly unconstrained due to the unknown depth, texture, and albedo of the scene.
%Illumination estimation techniques that use hand crafted lighting features and illumination priors tend not to generalize well across single-view, noisy outdoor datasets~\cite{}. 
Recent work applying deep learning techniques to sun location estimation have significantly higher performance than handcrafted features~\cite{peretroukhin2017reducing}. In fact, convolutional neural networks trained for sun location prediction have been shown to learn feature representations that activate to and detect lighting cues that are similar to those that humans use to estimate the sun direction, e.g., shadow regions, bright regions, etc~\cite{peretroukhin2017reducing,ma2017find}.
Furthermore, using sun location prediction as an additional loss has been shown to improve performance for visual tracking and odometry~\cite{peretroukhin2017reducing,ma2017find,clement2016improving}. 
The proposed method uses the same intuition by constructing a sun estimation loss that helps transfer illumination-specific information into the input RGB image. 
%So in general, illumination estimation from a single outdoor image is very a challenging issue. 
%
%%%%%%%%%%%%%%%%%%%%%%%%%%%%%%%%%%%%%%%%%%%%%%%%%%%%%%%%%%%%%%%%%%%%%%%%%%%%%%%%%%%%%%%%%%%%%%%%%
~\vspace{-2mm}
\subsection{Single and Multi-Image Relighting for real data}
The goal of image relighting is to change the existing illumination condition captured in an image or set of images to a target illumination condition. 
At their core, image relighting techniques attempt to model the light transport function, which is composed of a model of scene lighting and the BRDF (bidirectional reflectance density function) that describes the material properties throughout the scene~\cite{sun2007interactive,xu2018deep}. 
%Since the light transport function already combines all the light-scene interactions, these methods can reproduce highly photorealistic lighting effects that are difficult to reconstruct and render otherwise. Current state-of-the-art relighting methods utilize neural networks, which are incredibly powerful and successful at modeling highly nonlinear functions with a large number of dimensions, such as the BRDF and light transport function. 
However, the majority of deep relighting methods require multiple images (anywhere between 5-1000 depending on the image relighting method) of the same scene under different lighting conditions to generate an accurate estimate of the lighting function ~\cite{xu2018deep,philip2019multi}. Other methods require material labels of objects in the scene~\cite{alhaija2018geometric}. Neither of these types of datasets are easy to obtain for outdoor driving scenarios. In contrast, the proposed relighting method is self-supervised, and designed to operate on inputs that are readily available for such datasets.

%% jin2019single material guided single image relighting -- most similar to ours, NEED TO DISCUSS
%The relighting technique most similar to the proposed method is presented in [~\ref]. It is a single image relighting method that 

%%%%%%%%%%%%%%%%%%%%%%%%%%%%%%%%%%%%%%%%%%%%%%%%%%%%%%%%%%%%%%%%%%%%%%%%%%%%%%%%%%%%%%%%%%%%%%%%%
~\vspace{-4mm}
\subsection{Image to Image Translation for Illumination Adaptation}
%Recent years have witnessed rapid progress on generative adversarial networks (GANs) for image generation. 
%Image to Image translation methods have shown to be successfully used in day to night transformation. This suggests that these types of models can capture information relevant to scene lighting.
%(the networks have learned information relevant to scene lighting). 
%Thus, the relighting problem could be cast as a domain adaptation or image-to-image translation problem.

%Really what we want in Domain injection instead of domain adaptation, i.e, we want to inject more information into a given domain by increasing the dataset size and representation of visual effects in the dataset.
Many works have  cast the image relighting problem as a domain adaptation problem, and successfully used image to image translation methods to transfer images from the 'daytime' domain to the 'night time' domain~\cite{anoosheh2018night,sakkos2019illumination}. This suggests that these types of models may have an element to their design that can naturally capture information relevant to scene lighting. 
However, these methods become unstable and intractable when extending to multiple domains that would be necessary to capture transitions between multiple 'illumination condition/lighting' domains. There are several works that extend state-of-the-art domain to domain transfer methods to multi-domain to multi-domain transfer~\cite{anoosheh2018combogan, choi2018stargan}, but as far as we are aware, have not been applied to the particular problem of relighting.
%Such multi-domain to multi-domain methods would also require huge datasets in which time of day was labeled for each image, which is not feasible for real images.
%While there have been impressive successes in image-to-image translation models in projecting a given scene into the appearance/domain of another, e.g., day to night translation, they do not scale well to multiple domains, e.g., different times of day.
Furthermore, the outputs from image to image translation methods do not match the visual quality of real world data due to rendering artifacts introduced into the translated images that are not realistic or physically-based. Due to the highly unconstrained nature of these methods, it is not clear exactly what physical processes the model may be learning, and as a result there is no way to transfer specific subsets of the learned visual effects into images. 
In contrast, the proposed method avoids casting each light source location as a separate domain, and instead constrains the learned feature spaces of the Shadow Transfer framework through illumination-based losses and encoding networks. 

\begin{figure*}[t]
\begin{center}
\includegraphics[width=0.8\linewidth]{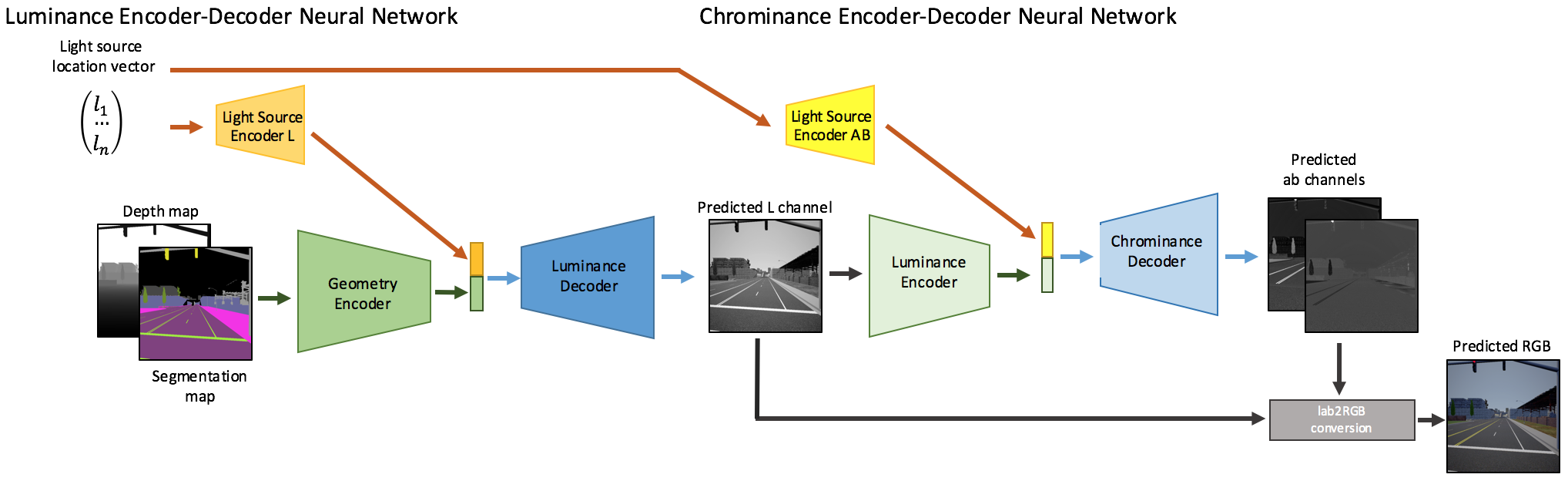}
\caption{The Shadow Transfer Network architecture. We break the domain transfer task into two steps: first, the luminace channel from the CIELab color space is predicted from the depth, semantic segmentation map, and light source location. Second, the chrominance channels, ab, are predicted from the predicted luminance channel and light source location. The luminance and chrominance predictions are concatenated together and transformed into RGB space to generate the final model output. 
%The stacked framework allows the predictions and visual information learned by the chrominance encoder-decoder to influence the learned representations of the luminance encoder-decoder, which better captures the coincidence of illumination effects between the L, a, and b channels. 
\vspace{-6mm}}
\label{fig:STnet}
\end{center}
\end{figure*}
%\vspace{0.5mm}
%
\section{Methods}
\label{sec:Methods}
%% include methods section
%%%%%%%%%%%%%%%%%%%%%
%% METHODS SECTION %%
%%%%%%%%%%%%%%%%%%%%%
%%
%The objective of the Shadow Transfer network is to learn the relationship between the scene geometry and the light source properties to generate accurate shading and coloration that is present in real RGB images. 
For autonomous vehicle driving datasets, typically we have access to RGB camera, depth sensor and GPS output. Pretrained, state of the art segmentation networks can also be used to generate semantic maps from the RGB video input, which captures information regarding objectness and material groupings within the scene. The proposed Shadow Transfer network is designed to operate on these intermediate representations to perform image relighting. The complete architecture for the proposed Shadow Transfer network is given in Figure~\ref{fig:STnet}.
%Since we do not have labels for the shadows or other illumination features within each RGB image, we apply the intuition/similar techniques from inverse graphics and intrinsic image decomposition towards constraining the learned/transferred features to be solely illumination effects in image to image translation frameworks. 
We define the scene lighting conditions as the light source location in the scene, which for outdoor images would be the sun azimuth and zenith angles. We choose this parameterization because altering the light source location in an image can simultaneously changes the illumination effects that have shown to be problematic for driving tasks, specifically shadow distributions, color temperature and brightness. 
%For the Shadow Transfer framework, we apply the intuition from inverse graphics and intrinsic image decomposition towards constraining the learned/transferred features in image to image translation frameworks. We cast the 
%Since we do not have labels for the shadows or other illumination features within each RGB image, we start with image reconstruction networks, specifically encoder-decoder frameworks.
 
%So our alternative solution to full illumination estimation is to focus upon a specific illumination cue (ie, constraining the image based relighting problem). 
%The accurate recognition of shadow area (i.e., shadow detection) provides adequate clues about the light sources, illumination conditions, object shapes and geometry information. 
%Shadows can either improve visual perception and understanding of a scene by adding important shape/orientation information about an object or cues about lighting/scene geometry, or they can add confusion, depending on whether they are modeled or ignored for a given task or taken into account based on the definition of the loss function.
%We choose to focus on altering the distribution of cast shadows and shading throughout the scene.

\subsection{Shadow Transfer Architecture}
%Since we do not have labels for the shadows or other illumination features within each RGB image, we start with image reconstruction networks, specifically encoder-decoder frameworks.
We adopt an framework that consists of two stacked encoder-decoder neural networks to perform two tasks: first, luminance prediction, and then second chrominance prediction. We choose to use the channels of CIELab color space representation for both tasks due to its success in shadow edge detection tasks~\cite{lalonde2012estimating,khan2005evaluation}.\\
%~\vspace{-1mm}
\textit{Luminance Prediction Network}\\
For luminance prediction, we use a U-net~\cite{ronneberger2015u} inspired encoder-decoder neural network to predict the luminance channel of the ground truth RGB image converted into CIELab space. As shown in the top of Figure~\ref{fig:STnet}, this network is comprised of three sub-networks: a geometry encoder that takes in the concatenated semantic segmentation and depth maps, a light source location encoder that takes in the location of the light source in the scene and projects it into the geometry encoder's latent space, and a luminance decoder that operates on the concatenated light source latent vector and geometry latent vector.\\
%~\vspace{-1mm}
\textit{Chrominance Prediction Network}\\
For chrominance prediction, we use a similar framework: a U-net inspired encoder-decoder framework that is comprised of a luminance encoder that takes in the predicted luminance channel, a light source location encoder that takes in the location of the light source in the scene and projects it into the luminance encoder's latent space, and a chrominance decoder that operates on the concatenated light source latent vector and the luminance latent vector to generate the chrominance channel predictions. Note that this is a similar process to the unsupervised colorization method proposed in~\cite{zhang2016colorful}. \\
%As shown in the bottom of Figure~\ref{fig:STnet}, this network is comprised of a light source location encoder (separate from the first), a luminance encoder that takes in the predicted Luminance channel, and a chrominance decoder that operates on the concatenated lightsource latent vector and luminance latent vector.
%
%~\vspace{-1mm}
\textit{Illumination-constrained latent space learning}\\
In classic image-to-image translation networks, an encoder neural network compresses and distills visual information into a latent space representation, which a decoder network then projects into an image. Ideally, each dimension (or subset of dimensions) of the latent space representation corresponds to a particular visual feature within an image. Thus, altering the values of the latent dimension alters the visual feature in the output image. However, in practice the learned latent spaces of image-to-image translation networks trained on complex scenes are not interpretable or necessarily smooth and easy to sample~\cite{worrall2017interpretable}.
%the magnitude and presence of a particular visual feature within an image, and thus can be used to control it. 
%To transfer only illumination effects, we wish to constrain the learned latent space such that we can externally define and control the lighting condition we desire in the input RGB image. 
To transfer only illumination effects, we proposed to externally enforce a subset of the latent space to correspond to light source location, similar to ~\cite{worrall2017interpretable}. This is achieved by injecting light source information into the luminance and chrominance latent spaces using the two separate light source encoders that are described above.
This design choice allows the two networks to learn which light source locations correspond to the shadow/shading distributions (based upon the extracted geometric information) as well as coloration that they induce in real images. \\
%~\vspace{-2mm}
\textit{Stacking Encoders for Multitask learning}\\
The motivation behind separating the tasks of luminance and chrominance prediction into a feed-forward multi-task framework stems from its success of in joint shadow detection and removal task~\cite{wang2018stacked,philip2019multi}. Modularizing and stacking the two tasks so that the output of one task is used as input to the other allows each network to focus on a single task at a time, reducing the complexity of the information that needs to be learned. Since the prediction occurs in two different stages, it allows each network to share mutual improvements through forward/backward information flows. 
This multitask framework specifically lends itself to luminance and chrominance prediction, primarily due to the co-incident nature of shadow and reflectance edges between the L and a channels, as well as the relationship between the `yellowness' of sunlight and respective `blueness' color of shadows within the b channel~\cite{khan2005evaluation}. This means that inaccurate predictions of the L channel would illicit inaccurate predictions of other illumination effects in the a and b channels, and this chrominance error would be back propagated into the luminance encoder-decoder network, providing a better training signal than luminance error alone.
%As observed in [11], while strong reflectance gradients are present in both the L and a channels, strong shadow gradients appear mainly in the L channel. 
%
\subsection{General Training procedure}
%In a single forward pass of the network, an image is first decomposed into LAB space
%\subsubsection{Loss Functions}
%Due to the difficulty in collecting and labeling autonomous vehicle datasets, 
The Shadow Transfer framework is designed to use a self-supervised training paradigm. 
%There are several loss functions used to train the Shadow Transfer framework. 
First, to ensure local semantic/structural consistency in each predicted image, we use apply the standard L1 loss to the predicted L channel and the predicted ab channels.
%\textbf{insert equations here}\\
This is a self-supervised loss because the ``labels", which are the ground truth CIELab channels, are obtained by merely computing the colorspace conversion on the RGB image. 
Since the family L norm losses are notorious for poorly reconstructing high frequency image information in local pixel neighborhoods, we use the standard perceptual and style loss proposed in~\cite{johnson2016perceptual} on the predicted RGB images, using the feature spaces of the VGG16~\cite{vgg16} architecture pre-trained on Imagenet~\cite{deng2009imagenet}.
%\textbf{insert equations here}\\
%Note that this is also essentially a self-supervised loss. 
However, none of these losses directly target the illumination effects that we want altered in the image.  The impressive results from~\cite{peretroukhin2017reducing} demonstrate that the neurons/layers in a CNN trained for sun location estimation in a single RGB image learned to activate in response to shadow and brightness regions in the image. Using this intuition, we pre-train a VGG16 network (initialized with weights pre-trained for scene classification on the Places-365 dataset) to perform sun location estimation on our ground truth RGB images. We use an L2 loss to train this network using the ground truth light source locations. We refer to this feature loss network as SunEst-CNN.
We then fix the weights of this SunEst-CNN, and during the training of the Shadow Transfer neural network, use its 'illumination' feature spaces to calculate an illumination-focused perceptual loss on the predicted RGB images. This is also a self-supervised loss, because it only requires the calculation of the light source location from the GPS and timestamps of the training dataset.
The total loss use to train the Shadow Transfer network is the sum of these four losses. 
%The sum of these losses are backpropagated through the network for each batch of RGB, depth, semantic segmentation maps and light source locations.
%\textbf{insert equations here}\\

%
%%  how to train the network diagram figure
%\begin{figure*}[ht]
%\begin{center}
%\includegraphics[width=1.0\linewidth]{figures/shadowtransfertraining_draft2.png}
%\caption{Shadow Transfer Network Training Paradigm}
%\label{fig:STnet_training}
%\end{center}
%\end{figure*}
%
%%%%%%%%%%%%%%%%%%%%%%%%%%%%%%%
%
%%%%%%%%% EXPERIMENTS TEXT %%%%%%%%%
\section{Experiments}
\label{sec:Experiments}
%% CARLA QUALITATIVE PROP METHOD VS BASELINES
\begin{figure*}
\begin{center}
\includegraphics[width=0.8\linewidth]{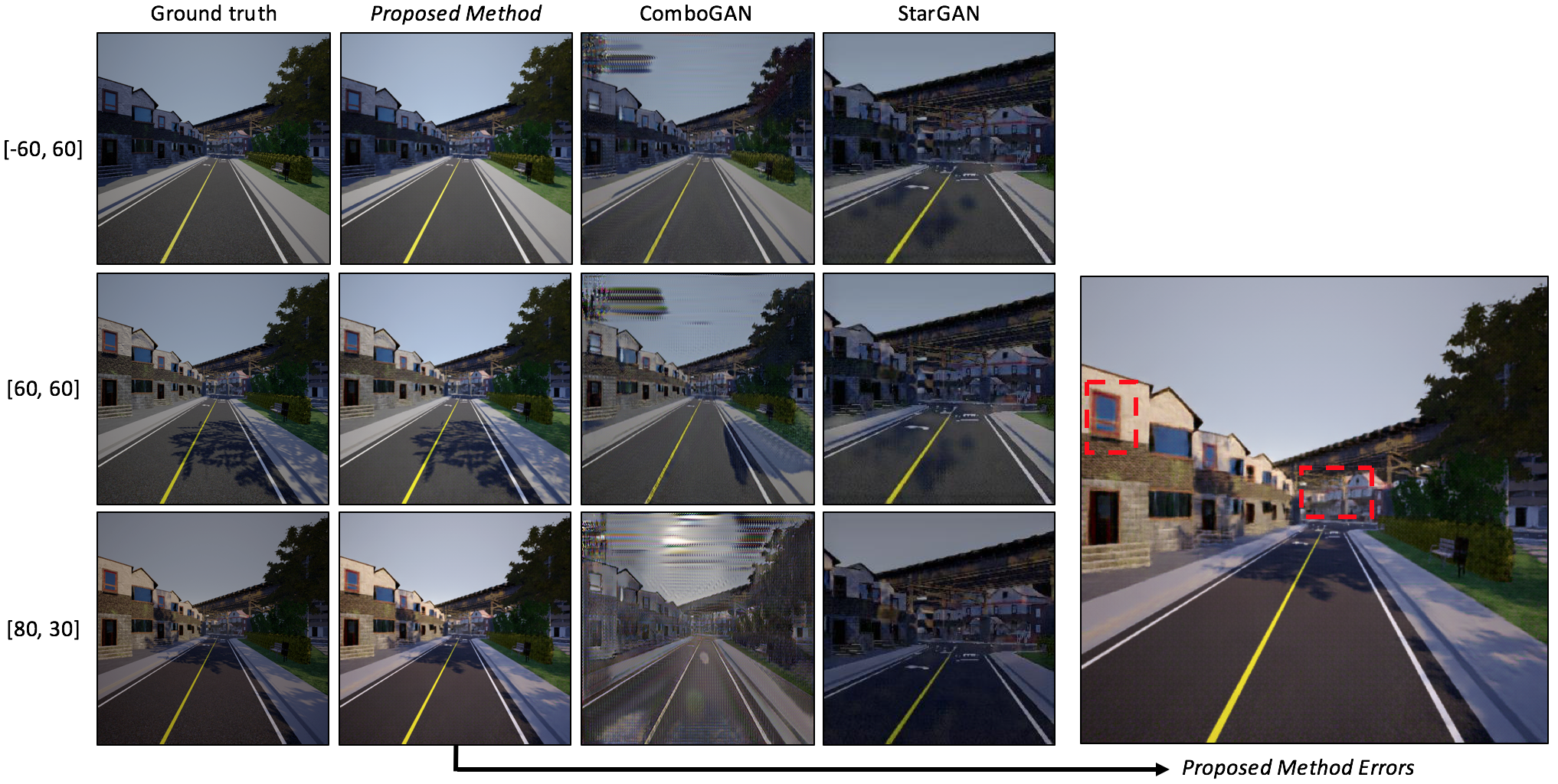}
\caption{This figure is best viewed in color in the web version. The above are example outputs from the proposed Shadow Transfer method (second column) in comparison to the state-of-the art baselines, ComboGAN (third column) and StarGAN (fourth column). Each row is the generated output image for a specified scene and light source location from the test set, given as a two element sun azimuth and zenith vector.}
\label{fig:carla-qualitative-baseline-comparisons}
\end{center}
\end{figure*}

%% TABLE: carla quantitative comparisons
\begin{table*}[]
\setlength{\tabcolsep}{3pt} %% default is 6pt
\centering
\begin{tabular}{|c|l|lccccccccc|}
\hline
\multicolumn{3}{|c|}{Model Type} & 
\multicolumn{1}{c|}{\begin{tabular}[c]{@{}c@{}}$\mu_{MSSIM}$\\ {[}-60.0, 60.0{]}\end{tabular}} & 
\multicolumn{1}{c|}{\begin{tabular}[c]{@{}c@{}}$\mu_{MSSIM}$\\ {[}80.0, 30.0{]}\end{tabular}} & 
\multicolumn{1}{c|}{\begin{tabular}[c]{@{}c@{}}$\mu_{MSSIM}$\\ {[}-80.0, 30.0{]}\end{tabular}} & 
\multicolumn{1}{c|}{\begin{tabular}[c]{@{}c@{}}$\mu_{MSSIM}$\\ {[}60.0, 60.0{]}\end{tabular}} & 
\multicolumn{1}{c|}{\begin{tabular}[c]{@{}c@{}}$\mu_{MSSIM}$\\ {[}0.0, 75.0{]}\end{tabular}} & 
\multicolumn{1}{c|}{\begin{tabular}[c]{@{}c@{}}$\mu_{MSSIM}$\\ {[}-95.0, 10.0{]}\end{tabular}} & 
\multicolumn{1}{c|}{\begin{tabular}[c]{@{}c@{}}$\mu_{MSSIM}$\\ {[}-15.0, 80.0{]}\end{tabular}} & 
\multicolumn{1}{c|}{\begin{tabular}[c]{@{}c@{}}$\mu_{MSSIM}$\\ {[}15.0, 80.0{]}\end{tabular}} & 
\begin{tabular}[c]{@{}c@{}}$\mu_{MSSIM}$\\ {[}95.0, 10.0{]}\end{tabular} \\ \hline
\multicolumn{3}{|c|}{\begin{tabular}[c]{@{}c@{}}\textit{Proposed}\\ \textit{Method}\end{tabular}} &\textbf{0.770}&  \textbf{0.704}&  \textbf{0.819}&  \textbf{0.721}&  \textbf{0.770}&  \textbf{0.803}&  \textbf{0.773}&  \textbf{0.769}& \textbf{0.810}  \\
\multicolumn{3}{|c|}{ComboGAN~\ref{}} &0.662 &0.602 &0.657  &0.674  &0.780  &0.721  &0.768  &0.766  &0.699 \\
\multicolumn{3}{|c|}{StarGAN~\ref{}} & 0.334 & 0.329 &0.339  & 0.292 & 0.330 & 0.375 &  0.327& 0.330 & 0.358 \\ \hline
\end{tabular}
\label{table:baselines_ssim}
\caption{Comparisons to state of the art multi-domain to multi-domain transfer methods}
\end{table*}
%% TABLE: carla ablation

% Please add the following required packages to your document preamble:
% \usepackage{multirow}
\begin{table*}[]
\setlength{\tabcolsep}{3pt} %% default is 6pt
\centering
\begin{tabular}{|cccccccccccc|}
\hline
\multicolumn{3}{|c|}{\begin{tabular}[c]{@{}c@{}}Shadow Transfer \\ Net Components\end{tabular}} & \multicolumn{1}{c|}{\multirow{2}{*}{\begin{tabular}[c]{@{}c@{}}$\mu_{MSSIM}$\\ {[}-60.0, 60.0{]}\end{tabular}}} & \multicolumn{1}{c|}{\multirow{2}{*}{\begin{tabular}[c]{@{}c@{}}$\mu_{MSSIM}$\\ {[}80.0, 30.0{]}\end{tabular}}} & \multicolumn{1}{c|}{\multirow{2}{*}{\begin{tabular}[c]{@{}c@{}}$\mu_{MSSIM}$\\ {[}-80.0, 30.0{]}\end{tabular}}} & \multicolumn{1}{c|}{\multirow{2}{*}{\begin{tabular}[c]{@{}c@{}}$\mu_{MSSIM}$\\ {[}60.0, 60.0{]}\end{tabular}}} & \multicolumn{1}{c|}{\multirow{2}{*}{\begin{tabular}[c]{@{}c@{}}$\mu_{MSSIM}$\\ {[}0.0, 75.0{]}\end{tabular}}} & \multicolumn{1}{c|}{\multirow{2}{*}{\begin{tabular}[c]{@{}c@{}}$\mu_{MSSIM}$\\ {[}-95.0, 10.0{]}\end{tabular}}} & \multicolumn{1}{c|}{\multirow{2}{*}{\begin{tabular}[c]{@{}c@{}}$\mu_{MSSIM}$\\ {[}-15.0, 80.0{]}\end{tabular}}} & \multicolumn{1}{c|}{\multirow{2}{*}{\begin{tabular}[c]{@{}c@{}}$\mu_{MSSIM}$\\ {[}15.0, 80.0{]}\end{tabular}}} & \multirow{2}{*}{\begin{tabular}[c]{@{}c@{}}$\mu_{MSSIM}$\\ {[}95.0, 10.0{]}\end{tabular}} \\ \cline{1-3}
\begin{tabular}[c]{@{}c@{}}Depth \\ Input\end{tabular} & \begin{tabular}[c]{@{}c@{}}Sem. Seg. \\ Input\end{tabular} & \multicolumn{1}{c|}{\begin{tabular}[c]{@{}c@{}}Sun Est. \\ Loss\end{tabular}} & \multicolumn{1}{c|}{} & \multicolumn{1}{c|}{} & \multicolumn{1}{c|}{} & \multicolumn{1}{c|}{} & \multicolumn{1}{c|}{} & \multicolumn{1}{c|}{} & \multicolumn{1}{c|}{} & \multicolumn{1}{c|}{} &  \\ \hline
 \checkmark & \checkmark &            &  \textbf{0.776}&  \textbf{0.707}&  \textbf{0.828}&  \textbf{0.732}&  \textbf{0.780}&  \textbf{0.810}&  \textbf{0.782}&  \textbf{0.780}& \textbf{0.817}\\
 \checkmark & \checkmark & \checkmark &  0.770&  0.704&  0.819&  0.721&  0.770&  0.803&  0.773&  0.769& 0.810 \\
 \checkmark &            & \checkmark &  0.721&  0.673&  0.760&  0.686&  0.721&  0.760&  0.724&  0.719& 0.756  \\
            & \checkmark & \checkmark &  0.762&  0.686&  0.803&  0.714&  0.763&  0.792&  0.765& 0.762 & 0.799 \\ \hline
\end{tabular}
\label{table:ablation_carla}
\caption{Ablation experiments of different inputs and network components for \textit{CARLA-sun}. The descriptor `Sun Est. Loss' indicates if the Shadow Transfer network was trained with the novel Sun-CNN feature loss. The descriptor `Depth Input' indicates if the Depth map was used as input. Similarly, the descriptor `Sem. Seg. Input' indicates if the semantic segmentation map was used as input. }
\end{table*}

%% CARLA MUNIT
%\begin{figure*}
%\begin{center}
%\includegraphics[width=1.0\linewidth]
%\includegraphics[width=1.0\linewidth]{figures/munit_on_carla.png}
%\caption{This figure is best viewed in color in the web version. The above are example outputs from transforming the same geometry representation of a scene into MUNIT with differing samples of style. While MUNIT is able to capture elements of illumination, such as color temperature, it does not realistically capture effects such as shadows. It also introduces in visual artifacts that are not present in real images, e.g., swatches of color in the sky and reflections on the road asphalt.}
%\label{fig:carla-munit}
%\end{center}
%\end{figure*}

%% KITTI ABLATION
\begin{figure*}
\begin{center}
\includegraphics[width=0.9\linewidth]{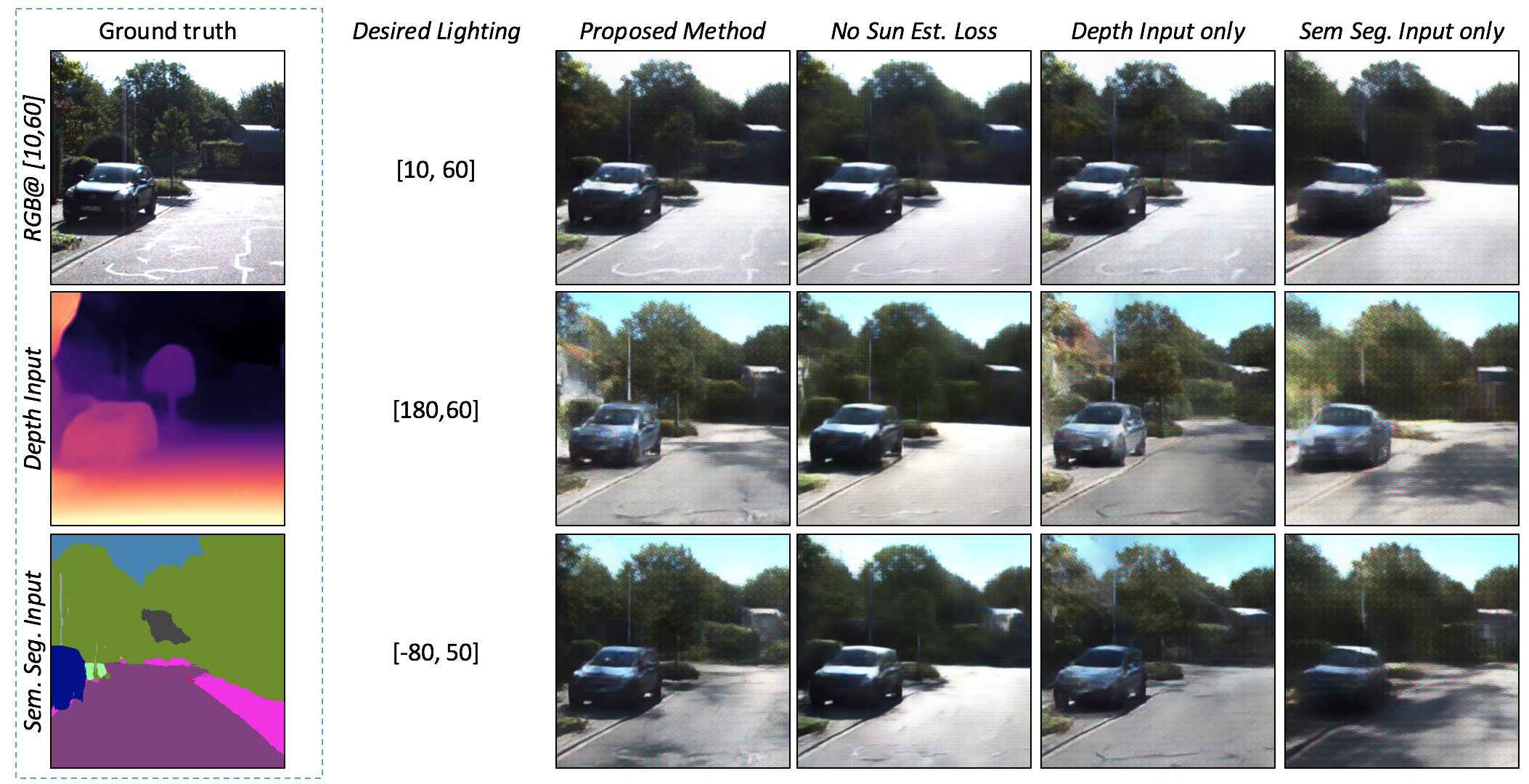}
\caption{This figure is best viewed in color in the web version of the paper. Shown above is the set of qualitative ablation experiments on the \textit{KITTI-sun} dataset. The original RGB image, Depth and Semantic Segmentation map inputs are provided in the first column in the blue dashed box. In the second column are the injected light source locations used to relight the original RGB image. The third through sixth columns are the outputs of the Shadow Transfer network when trained with either the full proposed model, the proposed model with no SunEst-CNN feature loss, the proposed model with only Depth input, and the proposed model trained only with Semantic segmentation input, respectively.}
\label{fig:kitti-ablation}
~\vspace{-4mm}
\end{center}
\end{figure*}

%%%%%%%%%%%%%%%%%
%% EXPERIMENTS %%
%%%%%%%%%%%%%%%%%
%
To validate the efficacy of our method, we present qualitative and quantitative evaluations on both real and synthetic datasets. The proposed Shadow Transfer network is most similar in design and goal to Multi-Domain to Multi-Domain transfer methods. Therefore, we compare the performance of the proposed method to the two state-of-the-art models StarGAN~\cite{choi2018stargan} and ComboGAN~\cite{anoosheh2018combogan}.

Since our goal is to generate images taken under lighting conditions/sun positions that don't necessarily exist in a given dataset, we cannot use quantitative metrics to validate the quality of generated images against ground truth for real image datasets. 
Therefore, we have designed a synthetic driving dataset using the CARLA rendering engine~\cite{dosovitskiy2017carla}, which we refer to as \textit{CARLA-sun}. Unlike real data, this dataset has a fixed number of light source locations, and contain the same scene under each possible lighting condition. To generate the dataset, we perform the same 100 vehicle trajectories under different sun positions defined by the sun azimuth and zenith angles. There are 9 total sun postions that correspond to the sun location at different times of day. There are a total of 13647 RGB, depth and semantic segmentation pairs, see the first row of Figure~\ref{fig:carla-qualitative-baseline-comparisons} for examples of the ground truth images. We also generated a held out test dataset of a single vehicle trajectory consisting of 424 images. We perform all synthetic experiments on this held out set. 
To test the proposed method on real data, we present results using the KITTI raw dataset~\cite{kittiobject}. We use the same method as ~\cite{peretroukhin2017reducing} to generate the sun position labels for each image in the dataset. We discretize the possible sun locations by rounding the azimuth and zenith to the nearest ten, which helps improve the latent space learning.
To demonstrate the self-supervised nature of the method, we use a pretrained state of the art monocular depth estimation network, Monodepth~\cite{godard2017unsupervised} to generate depth labels for each real image. Similarly we use the state of the art semantic segmentation network DeepLab~\cite{chen2017deeplab} pretrained on Cityscapes~\cite{Cityscapes} to generate coarse semantic segmentation labels for each real images. We refer to this dataset as \textit{KITTI-sun}, it has a total of 22400 RGB, depth, semantic segmentation, sun location pairs. 
For the \textit{CARLA-sun} synthetic dataset, we present ablation experiments to determine which of the Shadow Transfer network components yields the highest output image quality relative to the ground truth images. 
%We use the SSIM metric, which is a formula is based on three comparison measurements between the samples of luminance, contrast, and structure~\ref{}.
For \textit{KITTI-sun}, we present qualitative examples to determine if the components contribute to the perceived visual quality of the output images. 
%
%Therefore, we have designed two synthetic datasets of varying complexity, one based upon the CornellBox paradigm~\ref{} which we refer to as \textif{Synthetic CornellBox}, and one using the CARLA rendering engine, which we refer to as \textif{CARLA-sun}. Each of these datasets have a small, fixed number of light source locations, and contain the same scene under each possible lighting condition. 
%The Synthetic CornellBox dataset is loosely inspired by the famous cornell box paradigm in the graphics community. There are 250 unique scenes that contain 3D objects (a combination of spheres, cubes and cones) placed upon a textured ground plane. There are 14 possible (x,y,z) locations for the light source, and each scene is rendered under each possible light source condition. All surfaces and objects are diffuse, and the camera remains fixed in each scene. The total dataset contains 3500 pairs of RGB, depth, and semantic segmentation images.
%For both the Synthetic CornellBox and CARLA-sun synthetic datasets, we present ablation experiments to determine which combination of inputs (i.e., depth and semantic segmentation) and losses yeild the highest output image quality relative to the ground truth images. We use the SSIM metric, which is a formula is based on three comparison measurements between the samples of luminance, contrast, and structure.
%
\subsection{Training parameters and regimes}
%\vspace{-4mm}
For both the synthetic and real datasets, we train the Shadow Transfer networks for 50 epochs with a learning rate of 2e-4 and a batch size of 2. To accommodate the requirement of the U-net encoder-decoder, the input images are resized to 512x512.
The SunEst-CNN networks are initialized using the weights of a VGG-16 network trained to perform scene classification on Places-365. They are trained for 20 epochs at a learning rate of 1e-5, batch size of 2. The input images are resized to 256x256 to accommodate the architecture requirements.
The two state of the art multi-domain to multi-domain transfer methods, ComboGAN and StarGAN, are trained on \textit{CARLA-sun} using the training hyperparameters given in the paper and respective github repositories. 
%For our one-to-many domain transfer baseline, MUNIT, we train it for 540k iterations on \textif{CARLA-sun} using the training hyperparameters defined in the paper. 
All networks were train on a single Titan X GPU.
%\subsubsection{KITTI-sun}
%We train the Shadow Transfer networks for 70 epochs with a learning rate of 2e-4 and a batch size of 2. To accomodate the requirement of the Unet encoder-decoder, the KITTI-sun images are resized to 512x512. The sun-cnn networks are initialized using the weights of a VGG-16 network trained to perform scene classification on Places-365. They are trained for 20 epochs at a learning rate of 1e-5, batch size of 2, and the CARLA-sun images are resized to 256x256 to accomodate the architecture requirements.
%All networks are train on a single Titan X GPU. 
%\subsection{Evaluation on Synthetic CornellBox}
%\textit{Qualitative Evaluation}\\
%\textit{Ablation Experiments}\\
%\input{tables/cornellbox_ablation.tex}

\subsection{Evaluation on CARLA-sun}
%\vspace{-4mm}
\subsubsection{Comparison to Multi-Domain to Multi-Domain adaptation methods}
%% qualitative comparisons
We present a qualitative, visual comparison between the proposed method and state-of-the-art methods in Figure~\ref{fig:carla-qualitative-baseline-comparisons}. 
ComboGAN is able to capture realistic lighting, specifically shadows and color temperature, for each light source condition. However, it introduces translation artifacts that degrade the overall photorealism of the output image. StarGAN is only able to realistically capture varying global color temperature between the different light source domains. In contrast, the proposed method is able to capture realistic shading, shadows and color temperature without introducing the same artifacts as ComboGAN. Note that the proposed method appears to not match the brightness of the scene correctly in comparison to the ground truth, and also produces blur artifacts, as shown in the right hand side of Figure~\ref{fig:carla-qualitative-baseline-comparisons}.
These observations are verified in our quantitative analysis of the perceived generated image quality in comparison to ground truth, presented for each light source location in Table~\ref{table:baselines_ssim}. For a held out test trajectory from the \textit{CARLA-sun} dataset, we calculate the average mean structural similarity metric (MSSIM) for each light source location for the proposed method and baselines. StarGAN has the worst perceptual realism. We suspect this is because it was designed to operate on simpler scenes, specifically faces. Our model outperforms ComboGAN for each light source. As observed in Figure~\ref{fig:carla-qualitative-baseline-comparisons}, we suspect that this is because ComboGAN induces the same visual artifacts that plague the majority of image to image translation models trained to transform from RGB to RGB spaces. In contrast, the proposed method transforms from a geometric representation to an RGB representation, which yields a better mapping between visual features of the two spaces. 
%
%\subsubsection{Comparison to One-to-Many Domain adaptation methods}
%\input{fig/munit_comparison.png}
%To compare our stacked task network to a single network that transforms from one domain to another, we train MUNIT to transform from the 'geometry domain' to the RGB domain. 
%In Figure~\ref{fig:carla-munit}, we show example outputs from a state-of-the-art one-to-many domain transfer method, MUNIT, trained to project from the geometry domain (the concatenated semantic segmentation map and depth map) to an RGB image. While MUNIT is able to capture global illumination effects like color, it cannot produce realistic shadowe effects, and also introduces artifacts into the image that degrade the photorealism.
%\vspace{-1mm}
\subsubsection{Ablation Experiments}
To evaluate each component of our network, we calculate the mean MSSIM between the prediction and ground truth \textit{CARLA-sun} image for each possible light source location on the same held out sequence used above. Interestingly, it appears that having both the depth and segmentation map as input is more important to the predicted image quality than the SunEst-CNN feature loss. This suggests that the lighting conditions within the CARLA gaming system are not complex enough to need the extra constraint of the SunEst-CNN loss to accurately model within the Shadow Transfer framework. 

\subsection{Evaluation on KITTI-sun}
%\subsubsection{Training parameters}
%We train the Shadow Transfer networks for 100 epochs with a learning rate of 2e-4 and a batch size of 2. To accomodate the requirement of the Unet encoder-decoder, the KITTI-sun images are resized to 512x512 while maintaining the original aspect ratio. The sun-cnn networks are initialized using the weights of a VGG-16 network trained to perform scene classification on Places-365. They are trained for 25 epochs at a learning rate of 1e-5, batch size of 2, and the CARLA-sun images are resized to 256x256 to accomodate the architecture requirements. %All networks are train on a single Titan X gpu.
%
Since we are missing ground truth scenes for the relight real images, in this section we present a qualitative evaluation of the relight image quality. 
%\subsubsection{Qualitative Evaluation of Image quality and Ablation Experiments}
A set of qualitative ablation experiments are presented in Figure~\ref{fig:kitti-ablation}. When comparing the relight images generated by the full model of the proposed method to the model trained with out the SunEst-CNN feature loss, we see that there is no illumination feature transfer. This highlights the importance of the illumination-based SunEst-CNN feature loss during the training procedure for real data. Both the depth-only and semantic segmentation-only models are noticeably more noisy and blurry. The segmentation only model appears to better capture the changing shadow distributions throughout the scene and illumination features in general, where as the depth-only model appears to better capture realistic pixel statistics. 
In general, we observe that the proposed Shadow Transfer network has difficulty capturing local, high frequency noise. We suspect that this is due to the coarseness of the input depth and semantic segmentation masks.
%%%%%%%%%%%%%%%%%%%%%%%%%%%%%%%%%%%%
%
%%%%%%%%% CONCLUSION TEXT %%%%%%%%%
\section{Discussion and Conclusions}
\label{sec:concl}
%In general, our results show that the proposed Sensor Transfer Network reduces the synthetic to real domain gap more effectively and more efficiently than domain randomization. 
%suggest that the photometric effects of the sensor model have a significant contribution to the domain shift observed between benchmark vehicle datasets. It is clear however, that the measure used as the representation of the image data in the sensor domain ultimately determines the quality of the augmented data, and thus the effectiveness of the sensor transfer network.
%Future work includes increasingly the complexity and realism of the Sensor Transfer augmentation pipeline by modeling other, different sensor effects, as well as implementing models that better capture the pixel statistics of real images, such as motion or defocus blur. Other avenues include investigating the impact of task performance and problem space on the sensor effect parameter selection, and evaluating how the proposed method impacts performance for training synthetic datasets rendered with various levels of photorealism. 
In this work, we proposed a single image relighting framework that is able to successfully transfer illumination effects for both synthetic and real images. Our results indicate that the proposed Shadow  Transfer framework generates more realistic images than the state-of-the-art multi-domain to multi-domain transfer methods. Future work would be to apply advancements in the super pixel literature to this architecture to improve the high frequency noise modeling in the generated images. Other avenues/extensions of this work would be to incorporate a weather and imaging pipeline models
into this framework to better capture all of the myriad influences of illumination that impact image appearance. 
%%%%%%%%%%%%%%%%%%%%%%%%%%%%%%%%%%%%%%%%%%%%%%%%%%%%%%%%%%%%%%%%%%%%%%%%%%%%%%%%%%%%%%%%%%%%%%%%%%%%%%%%%%%%%%%%%%
%\clearpage
%\bibliographystyle{IEEEtran}
%\bibliography{main}
% Generated by IEEEtran.bst, version: 1.14 (2015/08/26)

%\printbibliography
%%%%%%%%%%%%%%%%%%%%%%%%%%%%%%%%%%%%%%%%%%%%%%%%%%%%%%%%%%%%%%%%%%%%%%%%%%%%%%%%%%%%%%%%%%%%%%%%%%%%%%%%%%%%%%%%%%

\end{document}